\documentclass[conference]{IEEEtran}
\IEEEoverridecommandlockouts
\usepackage[T1]{fontenc}
\usepackage[utf8]{inputenc}
\usepackage{microtype}
\usepackage{amsmath}
\usepackage{graphicx}
\usepackage{booktabs}
\usepackage{array}
\usepackage[numbers,sort&compress]{natbib}
\usepackage{url}
\usepackage[hidelinks]{hyperref}

\title{LEGAL-UQA: Testing Constitutional Question-Answering Systems for Urdu Speakers with a Parallel English--Urdu Benchmark}

\author{
\IEEEauthorblockN{Faizan Faisal\textsuperscript{1,*},
Umair Yousaf\textsuperscript{2,*},
Muhammad Hassnain\textsuperscript{1},
Zeerak Babar\textsuperscript{1}}
\IEEEauthorblockA{\textsuperscript{1}University of California, Davis, Davis, CA, USA\\
fznfaisal@ucdavis.edu, mhassnain@ucdavis.edu, zebabar@ucdavis.edu}
\IEEEauthorblockA{\textsuperscript{2}University of Colorado Boulder, Boulder, CO, USA\\
umair.yousaf@colorado.edu}
\thanks{\textsuperscript{*}Equal contribution.}
}

\begin{document}
\maketitle

\begin{abstract}
Legal-information assistants could help Urdu speakers understand their
constitutional rights, but building reliable systems requires test data and
evaluation methods suited to the language. We present LEGAL-UQA, a parallel
English--Urdu benchmark of 619 question--answer pairs linked to passages
from Pakistan's Constitution. Each example includes the question, reference
answer, and source passage in both languages. We evaluate two components
separately: answering Urdu questions when the source passage is supplied,
and retrieving the passage from which each question was generated. We also
compare JEV, a decision model used as an automated judge, with human grades.
The findings show that model rankings depend on how answer similarity is
measured, while a simple word-matching method retrieves source passages
more effectively than the embedding methods tested. At the evaluated
cutoff, JEV agrees with human grades less often than a baseline that gives
every answer full credit. An audit also identifies cases where generated
references omit information supported by the source. Together, these
findings highlight the need to test retrieval and answering separately,
include simple retrieval baselines, and validate automated judges and
reference answers with human review when engineering legal QA software for
Urdu speakers.
\end{abstract}


\section{Introduction}
Software that helps people understand their legal rights must work in the
languages they use. Language barriers in Pakistani legal services motivate
accessible Urdu resources \citep{ahmad2020english}. Article 251 of the
Constitution identifies Urdu as the national language and provides for
continued official use of English pending arrangements for its replacement
\citep{constitution_of_pakistan_english}. AI-enabled assistants that answer
constitutional questions in Urdu could widen access to legal information,
but only if their behavior can be tested in Urdu. Urdu legal QA combines
limited task-specific data with the need to preserve institutions,
conditions, exceptions, and procedural details in answers.

From a software-engineering perspective, a legal QA assistant combines at
least two components: one finds a relevant source passage, and one writes an
answer from it. Testing such a system needs inputs with known sources,
expected outputs, and an \emph{oracle} that decides whether an output is
acceptable. For free-form Urdu answers, the oracle is the difficult part.
Word-overlap metrics measure similarity to a reference rather than
correctness, and model-based judges must be checked for the language and
task at hand \citep{fu-liu-2025-reliable}.

LEGAL-UQA provides such test data for questions about Pakistan's
Constitution. Each example contains a question, an answer, and the source
passage in both English and Urdu. We call each manually delimited
constitutional article entry an \emph{article chunk}. The questions imitate
layperson wording but do not come from real consultations. The reference
answers were generated and translated, not written as legal opinions.

We test the two components separately. In \emph{supplied-article QA}, a
model reads the Urdu question and its source passage, then writes an Urdu
answer. In \emph{article retrieval}, a method searches the stored passages
for the one from which a question was generated. A correct retrieval result
is not itself a correct answer; the separation locates whether a failure
concerns finding the source or answering from it. We ask three research
questions:
\begin{itemize}
\item \textbf{RQ1 (answering):} How do Urdu answer sets compare when the
source passage is supplied, under reference-similarity metrics and final
human grades?
\item \textbf{RQ2 (retrieval):} How often do word-matching and embedding
methods recover the originating passage?
\item \textbf{RQ3 (oracle):} Can a decision model stand in for human
grading as an automated test oracle for Urdu answers?
\end{itemize}

The contributions are:
\begin{enumerate}
\item A parallel dataset of 619 English--Urdu QA pairs linked to 305 article
chunks, with both language versions of each source passage.
\item A component-level comparison of nine Urdu answer sets on 124
questions, alongside article retrieval results from searches based on
numerical text representations and on word matching.
\item A test of a decision model's agreement with final human grades on
1,116 Urdu answers, together with an audit of dataset coverage, article
overlap between data splits, answer references, and uncertainty, and the
implications for engineering inclusive legal QA software.
\end{enumerate}

\section{Related Work}
\label{sec:related_work}

\subsection{Urdu QA and Legal QA Resources}
The UQA dataset translates SQuAD2.0, a reading-comprehension dataset, into
Urdu and preserves answer spans \citep{arif-etal-2024-uqa}. UrFQuAD-v1 asks
questions for which a system must find a relevant Wikipedia passage,
including questions with no answer \citep{shakeel-nawab-2025-open-domain}.
\citet{kazi-etal-2025-crossing} compare reading comprehension across Urdu
and English questions and contexts. These datasets establish Urdu QA
methods, but their passages do not test whether a model preserves the
offices, conditions, and exceptions in Pakistan's Constitution.

Legal QA work shows the value of linking an answer to its source. BSARD
links Belgian citizens' questions to statutory articles
\citep{louis-spanakis-2022-statutory}; LLeQA adds detailed answers and
paragraph-level evidence \citep{louis-etal-2024-interpretable}. GerLayQA
pairs German lay questions with lawyer answers and cited paragraphs
\citep{buttner-habernal-2024-answering}, while EQUALS links Chinese
consultation questions to answer spans in law \citep{EQUALS}. KoBLEX asks
legal scenario questions in Korean and English that can require several
provisions \citep{lee-etal-2025-koblex}, and LexCLiPR tests multilingual
queries against English European Court of Human Rights paragraphs
\citep{upadhya-t-y-s-s-2025-lexclipr}. These resources motivate our
question--answer--passage structure, but their questions and laws address
other jurisdictions.

For Pakistani law, \citet{zafar-etal-2025-courtroom} provide data for
finding names of people, organizations, and other entities in legal text.
Urdu Legal Instruct describes 1,067 QA pairs across family, criminal, and
religious law, but its public description does not specify paired English
records or a source passage for each question
\citep{proxima-2026-urdu-legal-instruct}. PakLegalQA covers 300 questions
about federal statutes and Lahore High Court judgments, with evidence
citations, unanswerable questions, and a 60-question Roman Urdu version
\citep{irshad-2026-grounded}. It studies citation support and when a system
should decline to answer across a wider range of law. LEGAL-UQA instead
focuses on Urdu-script constitutional passages and provides paired
English--Urdu questions, answers, and source passages.

\subsection{Evaluating AI-Generated Answers}
Software-engineering research has evaluated language-model answers in
other domain-constrained QA settings: \citet{babar-etal-2026-stackexchange}
evaluate open-source models on technical StackExchange questions, using
retrieval-augmented generation and fine-tuning. Such evaluations depend on a
way to decide whether an answer is acceptable. Word-overlap scores cannot
decide whether a free-form answer conveys the same meaning as a reference.
A model-based judge can make that decision, but its agreement with people
needs to be checked for the language and task at hand.
\citet{fu-liu-2025-reliable} find that model-based judgments vary across
languages, with weaker consistency in lower-resource settings. In this
lower-resource Urdu legal setting, we compare JEV, a decision model
configured to judge reference equivalence, with final human grades. Here,
\emph{decision model} denotes a model that returns structured judgments and
probabilities for predefined questions, enabling software to apply explicit
decision rules. TypeSafe calls JEV a \emph{System One model}; it returns
typed outputs directly without generating free-form text
\citep{almeida-2026-jev}.

\begin{figure*}[t]
\centering
\includegraphics[width=0.86\textwidth]{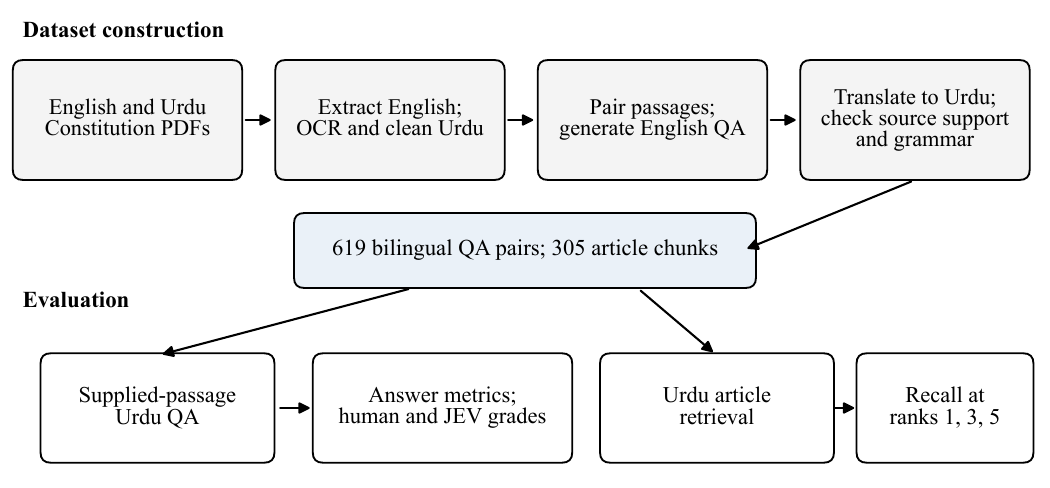}
\caption{Dataset construction and supplied-article QA evaluation. The diagram
shows the English and Urdu source documents, question generation, translation,
review, and answer evaluation. Answering models receive the question and
source passage; JEV also receives the reference and model answer.}
\label{fig:pipeline}
\end{figure*}

\section{Benchmark Construction}
\label{sec:dataset}
\subsection{Source Documents and Text Extraction}
The source documents are National Assembly English and Urdu versions of the
Constitution \citep{constitution_of_pakistan_english,constitution_of_pakistan_urdu}.
Both source editions state an amendment cutoff of 31 May 2018; their prefaces
are dated 31 May 2018 (English) and 12 July 2018 (Urdu). Our evaluations
concern the stored texts and make no claim about current law.

The English PDF has extractable text, while the Urdu PDF is image-based and
requires optical character recognition (OCR). YOLOv8 detects Urdu text-line
regions using the UrduDoc resources, and UTRNet recognizes text from the
cropped regions \citep{rahman_utrnet_2023}. Model-assisted cleanup with
GPT-4o uses the English article and OCR-derived Urdu text to correct
formatting, spelling, and sentence order. No independent transcription was
available to measure OCR error, so the cleaned Urdu text is not a verified
transcription.

After conversion to editable text, we manually segmented the English
document into 312 article-level passages, usually containing an article
heading and its clauses.
For example, Article 1's heading and three clauses form one chunk; Article
2's single sentence forms another. Some chunks contain only an
omitted-provision notice. Each passage received a unique identifier,
distinct from the Constitution's article number. The paired Urdu passages
were OCR-derived and cleaned, without an
independent check of exact article boundaries.

The 25th and 75th percentiles of English chunk lengths were used to separate
roughly the shortest quarter of chunks, the middle half, and the longest
quarter. In the saved categories, fewer than 45 whitespace-separated words
was small, 45--238 was medium, and 239 or more was large; the percentile
calculation and category assignment used different ways of counting words.
The prompt requested up to one, two, or three English QA pairs for small,
medium, and large chunks, respectively; these counts were not enforced
limits. The three categories contain 79, 157, and 76 chunks, average 31.05,
114.76, and 446.68 words, and contribute 79, 312, and 228 of the final QA
pairs.

\subsection{Question Generation, Translation, and Filtering}
An OpenAI GPT-4 model generated English questions and answers from each
article chunk, using a prompt that asks for layperson wording and forbids
article numbers in questions. Parsing the saved output yielded 621 English
QA pairs. Seven small chunks yielded no parsed pair, while seven other small
chunks yielded two despite the one-pair request. Two rows repeated an
English question already present, so removing them left 619 pairs with
distinct English questions.

Using GPT-4, we translated the English QA pairs with both language versions
of the article as context; the prompt asks for the Urdu constitutional
register. We checked the generated question--answer pairs against their
source passages for contextual validity and checked that both the Urdu
question and answer in each translated pair were grammatically correct.

We linked each English--Urdu QA pair to its bilingual source passage. The 619 final
pairs cover 305 article chunks. The seven chunks without pairs contain
omitted-provision notices or placeholder text. Each record contains English
and Urdu questions, answers, and passages, plus an internal chunk
identifier. The answers are free-form references, with no marked answer
spans. Automated checks confirmed complete bilingual records, unique
English questions, and consistent links between questions, passages, and
model responses, as well as coverage of all records by the data split.
These checks do not certify legal meaning. Figure~\ref{fig:pipeline} summarizes the
construction and the supplied-article QA evaluation.

\subsection{Coverage and Split}
Of the 305 chunks with questions, 67 have one QA pair, 162 have two, and 76
have three. Simple English-prefix counts find 195 yes/no forms, 175
``what'' questions, and 140 ``who'' questions; the remaining 109 use other
openings. These counts describe wording, not legal topics or reasoning
demands. English questions average 13.97 whitespace-separated words and
Urdu questions 15.64; English and Urdu answers average 25.93 and 28.71,
respectively.

The dataset was split at the \emph{question} level (80\% training, 20\%
validation; seed 25), giving 495 questions for model adaptation and 124 held
aside for evaluation. The split was not made by article: 95 chunks appear in
both sets. Thus 101 validation questions (81.45\%) concern a source passage
seen in training, although the English questions themselves do not repeat.
The results measure answers to new questions in a largely shared article
collection.

\section{Evaluation Design}
\label{sec:experiments}
\subsection{Answer Generators under Test}
Although the benchmark is bilingual, we evaluate answering only in Urdu;
retrieval is evaluated separately in Urdu and English. Each QA model reads
the Urdu question and its source article, then produces an Urdu answer.
Only mT5 is adapted on the LEGAL-UQA training split; the other models are
evaluated without task-specific fine-tuning on this dataset. We compare nine
models with one saved answer per question: Claude-3.5-Sonnet,
Gemini-1.5-Flash-002, Gemini-1.5-Pro-002, Mistral-Large, mT5,
GPT-5.4, DeepSeek-V4-Pro, Claude-Sonnet-5, and Gemma-4-E4B. The last four
runs requested temperature 0 and at most 1,024 output tokens, with reasoning
set to \texttt{none} (GPT-5.4), thinking disabled (DeepSeek-V4-Pro and
Claude-Sonnet-5), or no reasoning setting (Gemma-4-E4B). These four newer
models are also the focus of the additional uncertainty and normalization
analyses. They share an
instruction to answer only in Urdu script from the given context, but prompt
roles and wording differ among them.

For mT5, a multilingual text-generation model, we started with a checkpoint
trained on Urdu UQA \citep{arif-etal-2024-uqa,xue-etal-2021-mt5} and adapted
it for eight passes through the LEGAL-UQA training split (epochs), with
inputs limited to 512 tokens and target answers to 128. The same mT5
answers are used for automatic metrics, human grading, and JEV evaluation.

\subsection{Reference-Similarity Metrics}
\label{sec:automatic_metrics}
We first compare each answer with its single reference using token-overlap
F1 \citep{rajpurkar-etal-2016-squad}. Tokens are words separated by spaces.
Before counting, we lowercase text, remove ASCII punctuation and the words
``a'', ``an'', and ``the'', and normalize spaces. Let $P$ and $R$ be the
prediction and reference token lists, and let $c_P(w)$ and $c_R(w)$ count
occurrences of word $w$ in each. The per-answer score is
$2\sum_w\min(c_P(w),c_R(w))/(|P|+|R|)$; two empty answers score 1, while
only one empty answer scores 0. We report the mean on a 0--100 scale.

We also report METEOR \citep{banerjee-lavie-2005-meteor} on a 0--1 scale
and SacreBLEU \citep{post2018clarityreportingbleuscores} on a 0--100 scale.
Our METEOR implementation uses the Natural Language Toolkit's English-focused
word splitting, stemming, and WordNet synonym data, so it is not an
Urdu-specific semantic measure. SacreBLEU uses its \texttt{13a} word-splitting
scheme, preserves capitalization, and uses one reference with exponential
smoothing. As sensitivity checks, we calculate chrF
\citep{popovic-2015-chrf} over runs of 1--6 characters with $\beta=2$, and
Urdu-normalized token F1, which applies Unicode NFC, unifies common ``kaf''
and ``yeh'' forms and decimal digits, removes combining marks, tatweel, and
invisible formatting characters, and replaces punctuation with spaces. All
answer sets are scored with the same implementation. These metrics measure
reference similarity, not whether an answer is legally correct.

\subsection{Human Grading}
\label{sec:human_evaluation}
Human grading covered all 124 responses from each of nine model answer sets
(1,116 responses). Three proficient speakers of Urdu as their first
language participated. Two evaluators graded each response, and the third
resolved disagreements to determine the final grade; we use these final
adjudicated grades throughout. Graders saw the question, source passage,
reference answer, and model answer. Under our binary rubric, an answer
receives 1 only if it conveys the reference's meaning without omitting
details, allowing paraphrases. Correct additional information supported by
the source passage is not penalized; incorrect or partial answers receive
0. The evaluators had limited legal expertise.

\subsection{Retrieval Components}
\label{sec:retrieval_setup}
We search all 305 distinct source passages from the combined training and
validation data. A method succeeds when it retrieves the passage used to
generate the question. Recall@$k$ is the percentage of 124 questions for
which that passage appears in the first $k$ results. It measures recovery of
the recorded source, not whether other passages are legally relevant.

Two embedding models convert questions and passages into numerical vectors
and rank passages by cosine similarity; we report their results from saved
evaluation summaries. The lexical method BM25
\citep{robertson-zaragoza-2009-bm25} searches by matching words within each
language, using the stored Urdu and English questions separately. We apply
the Urdu normalization above, then split text at spaces, without stemming
or stop-word removal. BM25 uses $k_1=1.2$, $b=0.75$, and the
positive-smoothed inverse document frequency
$\log(1+(N-df+0.5)/(df+0.5))$. Each distinct query term counts once; ties
break by internal chunk identifier.

\subsection{Decision Model as Test Oracle}
\label{sec:judge_setup}
We used JEV (version 1.13.0) to compare each model answer with its reference.
Its input contains the Urdu question, source passage, generated reference,
and model answer. It returns a score
between 0 and 1 for reference equivalence; we count scores of at least 0.5
as positive. The judge prompt accepts paraphrases and
harmless additions but rejects contradictions, missing essential details,
and unsupported substantive additions. It calls the reference ``written by
a human expert'', although the references were generated and translated.
Although the source passage is supplied, the audit examines cases where
JEV's judgments appear to prioritize the reference over source-supported
information.

\section{Results}
\label{sec:results}

\begin{table*}[t]
\centering
\caption{Supplied-article Urdu answering on 124 validation questions per
answer set. F1, METEOR, and SacreBLEU measure similarity to generated
references; column maxima are bold. ``Human full'' is the full-credit rate
after the third reviewer resolved grading disagreements; ``JEV pos.'' is the
percentage JEV judged equivalent at the 0.5 cutoff; agreement and Cohen's
$\kappa$ compare JEV with the human grades.}
\label{tab:qa}
\begin{tabular}{lrrrrrrr}
\toprule
& \multicolumn{3}{c}{Reference similarity} & \multicolumn{4}{c}{Human grades and JEV judge} \\
\cmidrule(lr){2-4}\cmidrule(lr){5-8}
Answer set & F1 & METEOR & SacreBLEU & Human full (\%) & JEV pos.\ (\%) & Agreement (\%) & $\kappa$ \\
\midrule
Claude-3.5-Sonnet & 66.57 & 0.690 & 44.87 & 99.19 & 93.55 & 94.35 & 0.211 \\
Gemini-1.5-Flash-002 & 35.17 & 0.279 & 18.00 & 98.39 & 83.87 & 83.87 & 0.063 \\
Gemini-1.5-Pro-002 & 41.32 & 0.342 & 24.60 & 98.39 & 91.13 & 92.74 & 0.288 \\
Mistral-Large & 65.91 & 0.615 & 51.45 & 96.77 & 84.68 & 84.68 & 0.127 \\
mT5 & 62.22 & 0.605 & 42.81 & 74.19 & 65.32 & 79.84 & 0.527 \\
GPT-5.4 & \textbf{71.45} & 0.701 & 46.86 & 99.19 & 91.94 & 92.74 & 0.170 \\
DeepSeek-V4-Pro & 66.05 & 0.638 & \textbf{53.98} & 99.19 & 91.94 & 92.74 & 0.170 \\
Claude-Sonnet-5 & 68.71 & \textbf{0.703} & 45.88 & 99.19 & 91.94 & 92.74 & 0.170 \\
Gemma-4-E4B & 61.41 & 0.592 & 45.90 & 95.97 & 87.90 & 88.71 & 0.255 \\
\midrule
Pooled (nine human-graded sets) & & & & 95.61 & 86.92 & 89.16 & 0.336 \\
\bottomrule
\end{tabular}
\end{table*}

\subsection{RQ1: Answering with the Source Supplied}
\paragraph{Scores depend on the metric}
Table~\ref{tab:qa} shows that GPT-5.4 leads token F1 (71.45),
Claude-Sonnet-5 leads METEOR (0.703), and DeepSeek-V4-Pro leads SacreBLEU
(53.98). We estimate uncertainty for GPT-5.4, DeepSeek-V4-Pro,
Claude-Sonnet-5, and Gemma-4-E4B with 10,000 paired bootstrap resamples
(seed 25) of the 113 validation article chunks, where each sampled chunk
contributes all its questions. GPT-5.4's F1 is 2.75 points above
Claude-Sonnet-5, but the paired 95\% interval for this difference runs from
$-1.74$ to 7.38 points, so this sample does not establish a clear F1
difference between them. Its F1 leads over DeepSeek-V4-Pro (5.41, interval
1.18 to 9.90) and Gemma-4-E4B (10.05, interval 5.46 to 14.72) exclude zero.
These intervals reflect variation across sampled articles, not annotation
errors or repeated model runs, and are exploratory, with no adjustment for
testing several pairs.

\paragraph{Sensitivity analyses}
Urdu spelling normalization raises token F1 for GPT-5.4, DeepSeek-V4-Pro,
Claude-Sonnet-5, and Gemma-4-E4B by 1.40--1.91 points. Claude-Sonnet-5
leads chrF (69.86) and exact match (11.29\%), the share of answers identical
to their references under the token-F1 normalization; GPT-5.4 leads
Urdu-normalized F1 (73.36). Gemini Flash and Pro answers average 14.95 and
18.08 words, versus 28.88 for validation references. Their answers are
shorter, but these data do not show that length causes their lower overlap
scores.

\paragraph{Human grades}
Claude-3.5-Sonnet, GPT-5.4, DeepSeek-V4-Pro, and Claude-Sonnet-5 each
received full credit on 123 of 124 answers (99.19\%). The Gemini, Mistral,
and Gemma answer sets received full credit on 95.97--98.39\% of answers; the
mT5 set reached 74.19\%. Each of the three leading
reference-similarity models has the same human full-credit rate, so the
final grades do not separate them. One example makes an answer difference
concrete: asked who administers an oath, GPT-5.4 names the Chief Justice, as
the reference does, whereas Gemma-4-E4B answers ``Before the President,''
which states before whom the oath is taken rather than who administers it.
Their final human grades are 1 and 0; one example does not establish a
model-wide difference.

\begin{table}[t]
\centering
\caption{Originating-article retrieval for 124 validation questions over
305 passages (Recall@$k$, \%). Embedding scores come from saved Urdu
retrieval results; BM25 scores are calculated from the dataset.}
\label{tab:retrieval_comparison}
\footnotesize
\setlength{\tabcolsep}{2.5pt}
\begin{tabular}{llrrr}
\toprule
Method & Query language & R@1 (\%) & R@3 (\%) & R@5 (\%) \\
\midrule
text-embedding-3-large & Urdu & 53.23 & 72.58 & 77.42 \\
mistral-embed & Urdu & 29.03 & 45.97 & 52.42 \\
BM25 & English & 62.90 & 83.87 & 89.52 \\
BM25 & Urdu & 71.77 & 84.68 & 87.90 \\
\bottomrule
\end{tabular}

\end{table}

\subsection{RQ2: Retrieving the Source Passage}
\label{sec:retrieval}
As Table~\ref{tab:retrieval_comparison} shows, Urdu BM25 retrieves the
originating passage first for 71.77\% of questions and within the first five
for 87.90\%. The strongest embedding run, text-embedding-3-large, reaches
53.23\% and 77.42\%, respectively. English BM25 reaches 62.90\% at rank 1
and 89.52\% at rank 5. The mean reciprocal rank, the average of one divided
by the correct passage's rank, is 0.793 for Urdu BM25 and 0.751 for English
BM25. These are within-language searches over generated questions; the
results do not isolate a language effect or measure end-to-end QA.

\subsection{RQ3: The Judge as an Automated Oracle}
\label{sec:jev_results}
Across nine answer sets for the same 124 questions, JEV matched 995 of 1,116
final adjudicated human grades (89.16\%; Cohen's $\kappa=0.336$).
Cohen's $\kappa$ measures agreement after adjusting for agreement expected
by chance. Because
1,067 human grades were positive, always assigning full credit would agree
on 95.61\%. This baseline benefits from the predominance of full-credit
grades but detects none of the human-negative answers. JEV's ability to
identify those answers therefore matters alongside its overall agreement;
raw agreement alone does not establish reliable legal correctness.
Per answer set, agreement ranges from 79.84\% for the
mT5 set to 94.35\% for Claude-3.5-Sonnet (Table~\ref{tab:qa}); the human
full-credit rate is also the agreement of an always-positive decision for
that set, and it exceeds JEV's agreement for every set except mT5.

Of the judgments, 958 are positive under both JEV and the human grade, and
37 are negative under both. JEV marks 12 human-negative answers positive and
109 human-positive answers negative. It identifies 89.78\% of
human-positive answers and 75.51\% of human-negative answers. An
exploratory sweep selects a cutoff of 0.20 by maximizing pooled $\kappa$ on
these same labels (96.15\% agreement, $\kappa=0.578$), but this cutoff has
not been tested on held-out judgments. Answers to the same question recur
across models, so the 1,116 judgments are not independent observations. We
release the 121 JEV/human disagreements as review candidates; they are not
an exhaustive error set.

\subsection{Reference and Source Audit}
An answer can differ from the generated reference and still be supported by
the source. In one appeals question, the article permits an appeal when the
High Court takes over a lower-court case, tries it, and imposes a death or
life sentence. The reference omits this route. GPT-5.4 and DeepSeek-V4-Pro
include it and receive full human credit, but JEV rejects them despite
receiving the source passage. These judgments appear to follow the
incomplete reference rather than the supporting passage; the disagreement
does not show that the answers are wrong.

Translation can also change the reference itself. In a question defining
law, the Urdu reference limits an exclusion for fiscal and banking laws to
ten years, but the English reference omits the limit. These selected cases
call for expert bilingual review; they do not measure error frequency
across the dataset.

\section{Implications for Engineering Inclusive Legal QA Software}
\label{sec:implications}
\paragraph{Test components separately}
With the source passage supplied, final human grades give full credit to
95.97--99.19\% of answers from eight of the nine sets, whereas the best
retrieval method ranks the originating passage first for 71.77\% of
questions. These measures come from different tasks and do not predict
end-to-end accuracy, which needs its own test. They do show that a single
system-level score would not reveal which component failed, and that
component-level test sets with recorded sources make that diagnosis
possible.

\paragraph{Keep simple lexical baselines}
Urdu BM25 outperformed both embedding runs at rank 1 in our data.
Teams building Urdu assistants should not assume that embedding retrieval
is the stronger default; a word-matching baseline is cheap to include in
regression tests. Our questions were generated from the passages, so
independently written questions are needed to confirm this pattern for
real users' wording.

\paragraph{Validate automated oracles per language}
A raw agreement rate of 89.16\% looks high but falls below a trivial
always-full-credit baseline when most answers are acceptable. Before an
automated judge gates releases or ranks models for a new language, its
decisions should be compared with human grades and with that baseline,
reported with chance-corrected agreement and class balance, and its cutoff
selected on held-out questions or articles, keeping all answers to a
question in the same group. Supplying the source passage is insufficient
unless the judge also uses it to recognize correct information absent from
the reference.

\paragraph{Keep humans in the loop for references}
Generated and translated references can omit routes or add limits that the
other language lacks. Parallel English--Urdu records make such
discrepancies visible, but finding and correcting them requires bilingual
reviewers with legal expertise. Benchmark maintenance for under-served
languages should budget for that review rather than relying on generated
references alone.

\section{Threats to Validity}
\label{sec:threats}
\paragraph{Construct validity}
Reference similarity measures overlap with generated references, and METEOR
relies on English-focused resources. Our final human grades assess answer
completeness relative to the references and permit correct additions
supported by the source passages. The evaluators had limited legal
expertise; independent review by legal experts is needed to validate legal
correctness. References may omit or
mistranslate essential facts; selected examples do not establish how often
this occurs. Despite receiving source context, the judge can reject
source-supported answers that differ from the reference, and its prompt
inherits an erroneous expert-reference description.

\paragraph{Internal validity}
Prompt roles and reasoning settings differ across GPT-5.4,
DeepSeek-V4-Pro, Claude-Sonnet-5,
and Gemma-4-E4B. Each evaluated answer set contains one answer per question.
Metric intervals capture corpus sampling variability rather than variation
across repeated runs or differences in prompting. The 0.20 judge cutoff was
selected on the evaluated labels.

\paragraph{External validity}
The 619 QA pairs arise from one collection of constitutional passages and
generated question style. They support a pilot benchmark, not a claim that
this size suffices for broad legal training or robust model ranking.
Neither question prefixes nor length statistics establish semantic
diversity. The row-based split shares most validation articles with
training, limiting unseen-article generalization for the adapted model. OCR
cleanup and QA translation have not been independently measured for
preservation of legal meaning. Although both documents state the same
amendment cutoff, their provision-by-provision correspondence has not been
independently verified, and the resource does not cover later amendments.
Our results do not establish performance when a system must both find a
source passage and answer from it, reasoning across several legal
provisions, or suitability for legal advice.

\section{Ethics and Data Availability}
The resource is intended for research, including the evaluation of QA
software. Generated
or translated answers can misstate legal rights and procedures, and should
not replace qualified legal counsel. The construction uses public
constitutional documents and generated questions, with no collection of
private legal consultations. The English and Urdu contexts and references
must retain their provenance when redistributed. We use the MIT license for
our project code and dataset contributions; third-party constitutional
documents and model assets retain their own rights and terms, and this
designation does not grant rights the project does not hold. An analysis
manifest records dataset and response checksums, scoring library versions,
metric settings, and resampling settings.
The dataset, code, and analysis artifacts are available here:
\url{https://github.com/muhammad-hassnain/LEGAL-UQA}.

\section{Conclusion and Future Work}
LEGAL-UQA links English and Urdu questions and answers to constitutional
source passages, supporting separate tests of answering from a supplied
passage and finding that passage. The leading QA model depends on the text
similarity metric; Urdu BM25 retrieves the originating passage more often
than the reported embedding runs at rank 1. Selected omissions in reference
answers show that a source-supported answer can differ from its reference.
At the 0.5 cutoff, JEV's agreement with final human grades is below an
always-full-credit baseline, so decision models used as judges need per-language
validation before they serve as test oracles for Urdu answers.

Future work should independently review the legal content, split training
and evaluation by article, and compare answering in English and Urdu with
both supplied and retrieved passages. Independently written questions
should test conditions, exceptions, insufficient evidence, and when a
system should decline to answer. Formal methods have been applied to
statutory reasoning \citep{holzenberger-etal-2020-sara} and to abstract
notions such as software trust \citep{hassnain-etal-2026-cargo-sherlock};
similar encodings could strengthen this pipeline, for example by checking
answers against the source provision.

\bibliographystyle{IEEEtranN}
\bibliography{references}

\end{document}